\RequirePackage{amsmath}
\documentclass[runningheads]{llncs}
\usepackage[T1]{fontenc}
\usepackage{graphicx}
\usepackage{subcaption}
\usepackage{color}
\usepackage[dvipsnames]{xcolor}
\usepackage{multirow}
\usepackage{amssymb}
\usepackage{amsfonts}
\usepackage{algorithm2e}
\usepackage{bm}
\usepackage[right, pagewise, displaymath]{lineno}

\newcommand{\algoabbr}{{e}TOP}

\newcommand\Tstrut{\rule{0pt}{2ex}}         

\newcommand{\Nev}[1]{\MakeLowercase{}{#1}}
\newcommand{\Nm}[1]{\bm{#1}}

\newcommand{\Ng}[1]{\mathcal{#1}}
\DeclareMathAlphabet{\mathsfit}{\encodingdefault}{\sfdefault}{m}{sl}
\SetMathAlphabet{\mathsfit}{bold}{\encodingdefault}{\sfdefault}{bx}{n}
\newcommand{\tens}[1]{\bm{\mathsfit{#1}}}

\newcommand{\Nt}[1]{\tens{#1}}

\newcommand{\Ns}[1]{\mathbb{#1}}

\begin{document}

%
%
\title{\algoabbr: Early Termination of Pipelines for \\ Faster Training of AutoML System
\thanks{This work was done at Summer Internship at Nokia Bell Labs, NJ, USA}
}

\titlerunning{eTOP: Early Terminate of Pipelines}
%
\author{Haoxiang Zhang\inst{1} \and Juliana Freire\inst{1} \and Yash Garg\inst{2}}
\authorrunning{Zhang et al.}
\institute{
New York University, Brooklyn, NY, USA \\
\email{\{haoxiang.zhang, juliana.freire\}@nyu.edu} 
\and Nokia Bell Labs, Murray Hill, NJ \\
\email{yash.garg@nokia-bell-labs.com}
}
\maketitle              
\begin{abstract}
    Recent advancements in software and hardware technologies have enabled the use of AI/ML models in everyday applications has significantly improved the quality of service rendered. However, for a given application, finding the right AI/ML model is a complex and costly process, that involves the generation, training, and evaluation of multiple interlinked steps (called pipelines), such as data pre-processing, feature engineering, selection, and model tuning. These pipelines are complex (in structure) and costly (both in compute resource and time) to execute end-to-end, with a hyper-parameter associated with each step. AutoML systems automate the search of these hyper-parameters but are slow, as they rely on optimizing the pipelines' end output. We propose ``\textit{\underline{the \algoabbr~Framework}}'' which works on top of any AutoML system and decides whether or not to execute the pipeline to the end or terminate at an intermediate step. Experimental evaluation on \textit{26} benchmark datasets \footnote{OpenML - \url{https://www.openml.org/search?type=data}} and integration of \algoabbr with MLBox\footnote{MLBOX - \url{https://github.com/AxeldeRomblay/MLBox}} reduces the training time of the AutoML system upto \textbf{40x} than baseline MLBox. 
\end{abstract}
%
\section{Introduction} \label{sec:intro}
\begin{figure}[t]
    \centering
    \begin{subfigure}{0.49\linewidth}
        \centering
        \includegraphics[width = 0.95\linewidth]{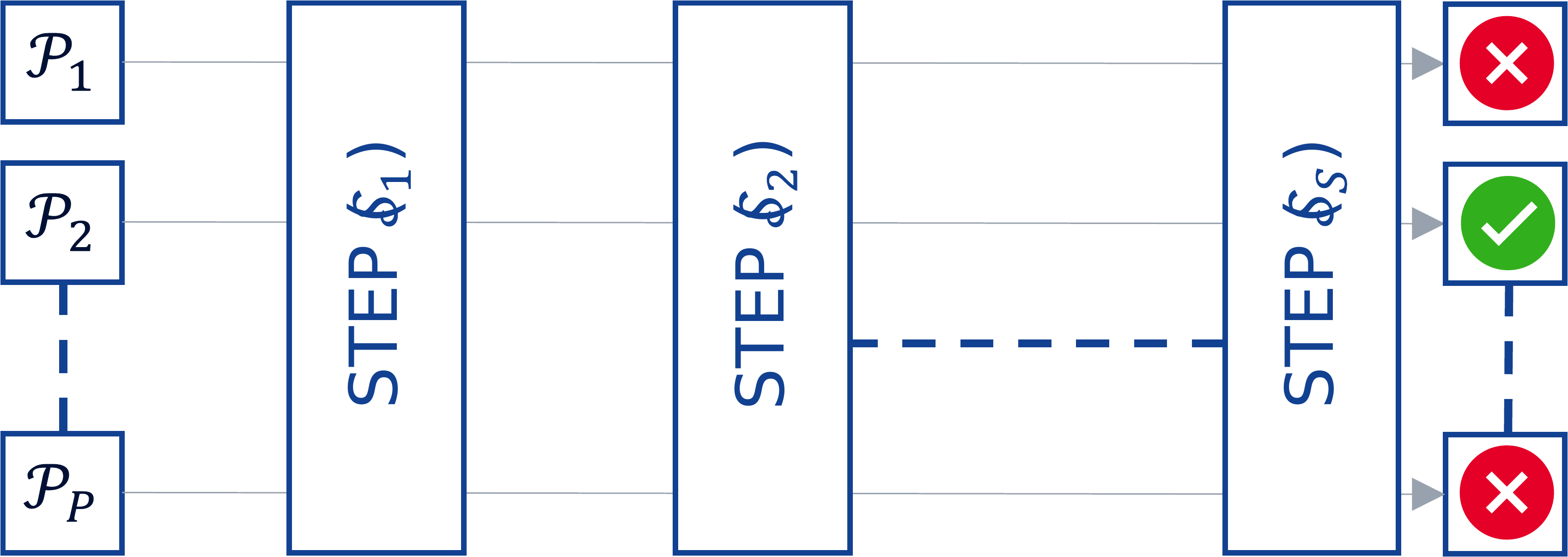}
        \caption{Standard AutoML System}
        \vspace{0.1cm}
        \label{fig:intro:automl:conventional}
    \end{subfigure}
    \begin{subfigure}{0.49\linewidth}
        \centering
        \includegraphics[width=0.95\linewidth]{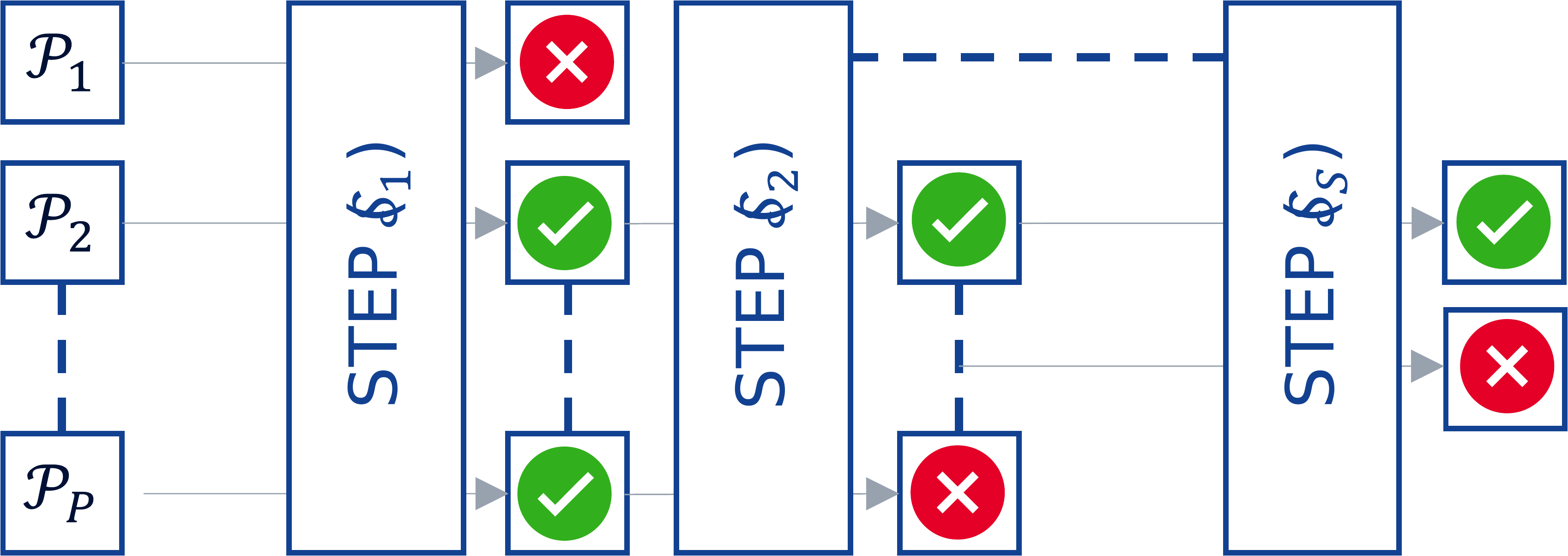}
        \caption{AutoML System powered with \algoabbr}
        \label{fig:intro:automl:etop}
    \end{subfigure}
    \caption{Architectural overview highlighting the critical difference in how pipelines are searched and executed in eTOP vs standard AutoML systems}
    \label{fig:intro:automl}
\end{figure}

Training classical machine learning models, such as Support Vector Machines (SVM) and decision tree-based models (DT), involves the optimization of multiple hyper-parameters, let alone deep models. Such optimization involves hand-crafted search space that results in 100s and 1000s of possible hyper-parameter configurations that must be considered and evaluated empirically. This process takes a considerable amount of computing time and resources. Many works have been proposed to address the hyper-parameter search, such as grid search \cite{larochelle2007empirical}, random search \cite{bergstra2012random}, and Bayesian optimization \cite{movckus1975bayesian}. More recently, \cite{garg2019racknet} focused on zero-shot hyper-parameter search for deep networks by performing pre-training analysis of the data.

Contemporary hyper-parameter search techniques (as shown in Figure \ref{fig:intro:automl:conventional}) have advanced to include not only model hyper-parameters but also hyper-parameters involved in machine learning pipelines (sequence in individual steps) outside the model training, such as data cleaning, imputation, and wrangling \cite{feurer2020auto,olson2016tpot,mlbox2018romblay}. Such techniques are now referred to as AutoML Systems ($\Ng{A}$) that automate, or minimize the need for human intervention during the search process. Furthermore, AutoML systems aim to automate the search of a large space of the Machine Learning (ML) pipeline. These AutoML systems typically work following an iterative heuristic, i.e. given a dataset ($\Ng{D}$) of interest, they first define a search space ($\Ng{S}$) and generate a set of corresponding pipelines ($\Ns{P} in {P_1, P_2,\dots,P_P}$). Each pipeline is further composed of interlinked sequential steps ($s$) that reflect different parameterized operations in the machine learning pipeline. In order to find the best hyper-parameter setting or pipeline, these pipelines are individually trained (end-to-end) and tested after the completion of the last step. 

\subsection{Key Limitations}
While the iterative search (such as grid search or random search) has shown success, however, the search space explodes when pipelines contain many steps with an even greater number of choices. This explosion in the number of candidate pipelines is prohibitive of compute time and cost of resources needed to support searching through large hyper-parameter search space.

The significant time cost of AutoML systems comes from evaluating the \textit{end-to-end} pipeline. As the quality of the pipeline is only measured after the completion of the last step. Time Constrained approaches have been proposed that limited the amount of time available for the AutoML system to search through the space, however, this came with a trade-off - balancing between time and pipeline accuracy \cite{lu2019hyper}. Thus the resulting pipeline was mostly sub-optimal.

\begin{figure*}
    \centering
    \includegraphics[width=1.0\linewidth]{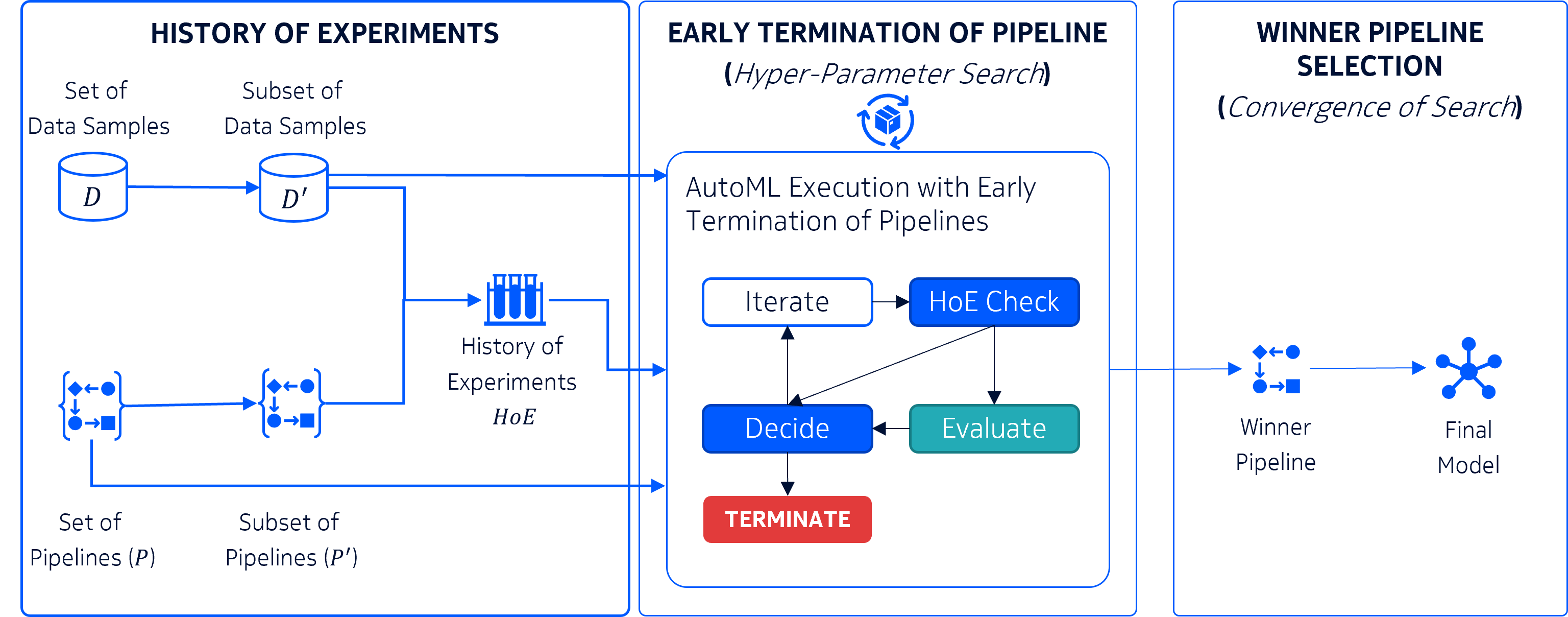}
    \caption{An Abstract overview of the proposed \algoabbr~framework, which is organized into 3 steps: first, generating a history of experiments, second, Hyper-Parameter Search with Early Terminate of Pipeline, and third, selection of winner pipeline. Details in Section \ref{sec:proposed_work}.}
    \label{fig:intro:system_overiew}
\end{figure*}

\subsection{Key Contributions of \algoabbr}\label{sec:intro:contribution}
To this end, we observe that one can achieve results that are ``\textit{near optimal} -- \textit{if not optimal}'' -- by intelligent determining the quality of a pipeline at an intermediary step in the search process and terminating the pipeline early as opposed to the after pipeline at the end of last step (as shown in Figure \ref{fig:intro:automl:etop}). We propose the \textbf{\underline{\algoabbr}}~Framework, an add-on to any existing AutoML systems that allow them to self-determine at an intermediate step with the pipeline to continue executing the pipeline or terminate early, instead of exploring the pipelines end-to-end (as shown in Figure \ref{fig:intro:system_overiew}). As a result, \algoabbr~can accelerate the AutoML pipeline search process and thus makes the training of AutoML systems faster. Described in detail in Section \ref{sec:proposed_work}.
%
%
%
%
%
%

The salient contributions of \algoabbr~are
\begin{itemize}
    \item \textbf{Pipeline Structure Agnostic}: \algoabbr~does not assume any fixed pipeline structure or sequential order. Thus able to accommodate pipelines of varying lengths and interlinks. This is possible due to the inherent design of \algoabbr,i.e. as it can be plugged after any individual step in the pipeline.
    \item \textbf{Low-Cost Pipeline Quality Measure}: \algoabbr~employs low-cost surrogate models that estimate the quality of pipelines at different steps, this allows for minimal overhead in comparison to the cost of training all pipelines end-to-end.
    \item \textbf{Intra-Pipeline Knowledge Sharing}: Furthermore, \algoabbr~self-determines pipeline termination by comparing its intermediate with the intermediate results from other pipelines.
\end{itemize}

%
%
%

\subsection{Organization}
The rest of this paper is organized as follows: Section \ref{sec:related_work} provides an overview of the existing state-of-the-art, followed by Section \ref{sec:proposed_work} which describes -- in detail -- the proposed \textbf{\underline{\algoabbr~Framework}}, Section \ref{sec:experiments} presents the extensive evaluation of the proposed work, and Section \ref{sec:conclusion} concludes.

\newpage 
\section{Related Work}
\label{sec:related_work}
The increase in the application of AI/ML models in everyday applications and critical applications such as predictive maintenance \cite{zaman2015recommender,garg_akyamac_lehman}, medical \cite{desai2021anatomization}, and emergency response \cite{ai2019intelligent} has increased the need for curating the models for specific applications. Classical machine learning models to state-of-the-art deep learning models are complex models that involve multiple hyper-parameters that require tuning, i.e. finding the right setting to train an accurate model for a given data/task. The process of finding these settings is called ``\textit{hyper-parameter search}'' and the set of all possible settings are referred to as ``\textit{search space}''.
\vspace{-0.15in}
\subsection{Hyper-Parameter Search Space}
Classical hyper-parameter search has been around for many decades, starting from grid search \cite{larochelle2007empirical}, and Bayesian Optimization \cite{movckus1975bayesian}. The former relies on principled greedy heuristics that search through each and every hyper-parameter choice in the hand-crafted search space defined by the expert. However, this grid-based search is a slow and costly process. To this end, the latter proposed to select starter hyper-parameter choices randomly and then iteratively search around the few set-performing ones. This helped reduce the need for exploring all hyper-parameter settings. 
\vspace{-0.15in}
\subsection{AutoML System}
As noted in Section \ref{sec:intro}, the present-day need for hyper-parameter search is no longer centered around models only but also around steps such as data pre-processing and engineering - to name a few. Thus, many AutotML systems were designed, such as MLBox \cite{mlbox2018romblay}, AutoSkLearn\cite{feurer2015efficient}, and TPOT \cite{olson2016tpot}, to address the increasing complexities of hyper-parameter search. The first step these AutoML took was to automate the process of searching and configuring the best machine learning pipeline (combination of hyper-parameter choices). This automation liberated the data scientists from the tedious and time-consuming process of hyper-parameter search. Despite the wider benefits of this automation, the large and complex search space still exists, which makes the AutoML system computationally expensive. Many AutoML systems have been focusing on optimizing the pipeline searching time. 
\vspace{-0.15in}
\subsection{Pipeline Optimization}
\cite{yakovlev2020oracle} performs the optimization at two stages: first (offline) to identify a collection of proxy models for different types of models, and second (online) to evaluate these proxy models with a reduced dataset and selecting the most accurate proxy model. 
\cite{feurer2020auto} proposed two-step optimization: first (offline) to acquire a set of accurate pipelines across multiple datasets; and second (online), given a new dataset, all pipelines in the portfolio are evaluated iteratively using Bayesian Optimization (BO). 
%
%
However, all these AutoML systems require a large amount of time for meta-learning to generate a reference of history experiments. Furthermore, the optimization is focused on inter-pipeline performance, i.e. the winner pipeline is selected based on their performance after it is executed end-2-end irrespective of the optimization strategy.


%
\newpage
\section{The \algoabbr~Framework}\label{sec:proposed_work}
\begin{table}[t]
    \centering
    \caption{Notations used in the paper}
    \label{tab:appendix:notation}
    \resizebox{\linewidth}{!}{
        {\setlength{\tabcolsep}{1em}
        \begin{tabular}{|r|l|r|l|r|l|}
            \hline
            \multicolumn{2}{|c|}{\textbf{General}} & \multicolumn{2}{|c|}{\textbf{Step}} & \multicolumn{2}{|c|}{\textbf{Pipeline}} \\ \hline
            \textbf{Notation} & \textbf{Definition} & \textbf{Notation} & \textbf{Definition} & \textbf{Notation} & \textbf{Definition} \\
            \hline
            $\Ng{A}$ & AutoML Systems & $\Ng{S}$ & Step Instance & $\Ng{P}$ & Pipeline Instance \\ 
            $\Ng{M}$ & AI/ML Model & $\Nev{s}$ & Step Iterator & $\Nev{p}$ & Pipeline Iterator \\
            $\Ng{M}^{s}$ & Surrogate Model & $\Ns{S}$ & Set of Steps & $\Ns{P}$ & Set of pipelines \\ 
            & & $\Nt{S}$ & Step Search Space & $\Nt{P}$ & Pipeline Search Space \\
            \hline
        \end{tabular}
        }
    }
\end{table}

In this section, we present (in detail) the proposed ``\underline{\textit{eTOP Framework}}'' that helps significantly reduce the AutoML training time while preserving (of at minimal trade-off) the end model accuracy. \algoabbr~operates with a simple assumption that not every pipeline ($\Ng{P}\in \Ns{P}$, here, $\Ns{P}$ is a set of pipelines) needs to be executed end-2-end (i.e. from start to end).
To be specific, \algoabbr~performs a step-wise conditional execution of AutoML s.t. the mediocre pipelines will be terminated at an earlier step without running the pipeline until the last step. To do so, \algoabbr~leverages early markers to decide if and when to the pipeline.

\subsection{Prerequisites}
Let's first begin by formally defining the key concepts used in this work:
\begin{definition}
    \label{def:proposed:step}
    \textbf{Step} ($\Ng{S}$) is an isolated computational operation in an AutoML Pipeline. Furthermore, a step can be further classified into two groups,
    \begin{itemize}
        \item \underline{\textit{data step}} ($\Ng{S}^{d}$) is a type of step that focuses on data operations, i.e. if take data in a specific format as input and output data in a new format or representation., s.t., $\Nm{X}^{new} \leftarrow \Ng{S}^d(\Nm{X}^{old})$.
        \item \underline{\textit{model step}} ($\Ng{S}^{m}$) is a type of step that focuses on learning a model for a specific task, such as classification or regression. This step takes data (in any form) and target variable as input ($\Nm{X}$) and output ($\Nm{Y}$) a trained model ($\Ng{M}$) and associated performance metric ($acc$), s.t. $\langle acc, \Ng{M} \rangle \leftarrow \Ng{S}^m(\Nm{X}, \Nm{Y})$.
    \end{itemize} 
\end{definition}
\begin{definition}
    \label{def:proposed:pipeline}
    \textbf{Pipeline} ($\Ng{P}$) is an interlinked set of steps, $\Ns{S}$, where steps execute in a cascade fashion, s.t. $\langle acc, \Ng{M} \rangle \leftarrow \Ng{P}(\Nm{X}, \Nm{Y} \vert \theta)$, here, $\theta$ represents the set of choices selected for different steps in the pipeline.
\end{definition}
\begin{definition}
    \label{def:proposed:surrogate}
    \textbf{Surrogate Model} ($\Ng{M}^{s}$) is a classic machine learning tree-based model used to approximate the intermediate quality of data, s.t. $acc \leftarrow  \Ng{M}^{s}(\Nm{X}, \Nm{Y})$. As noted in \cite{shwartz2022tabular}, tree-based models are known to outperform AI-based models on tabular datasets. Thus, we employ tree-based surrogates as our experimental evaluation is on classification tasks for tabular dataset\footnote{\algoabbr~does not tune the hyper-parameters of the surrogate models as the main goal is to limit the overhead of \algoabbr~and to maintain deterministic and invariable evaluation criteria}.
\end{definition}
\newpage
\noindent

Next, in this section, we present the key components (as shown in Figure \ref{fig:intro:system_overiew}) of the \algoabbr~framework that enable early termination of pipelines and consequently provide a significant boost in reducing of training time of AutoML Systems.

Given an AutoML system ($\Ng{A}$) and its pipeline search space ($\Nt{P}$), and a dataset ($\Nm{D} \leftarrow \langle \Nm{X}, \Nm{Y} \rangle$) for each $\Ng{A}$ has to search for a high performing pipeline ($\Ng{P}^{o}$). Our proposed strategy, first, generates a history of experiments (HoE) to establish a baseline of pipeline performance, followed by an iterative, but conditional, pipeline search while maximizing the intra-pipeline-step performance .
 
%
\vspace{-0.1cm}
\subsection{History of Experiments}
Motivated by the well-established concept of meta-learning, \algoabbr~aims to establish a baseline performance for the pipelines and the individual steps associated with the pipelines. Generating the history of experiments (HoE) involves the following steps - execute in a cascade fashion: 
\begin{enumerate}
    \item\textbf{ Data Sampling}: This step generates a subset of data instances and pipelines
    \begin{itemize}
        \item \textbf{Representative Stochastic Data Sampling}: This step performs a class-distribution-aware random sampling, i.e. this sampling maintains the class distribution in the sampled population that is identical to the original population. This heuristic serves two purposes: first, real-world data has class imbalance thus the sampled population will reflect the same., and second, random sampling doesn't assume or hold subconscious bias towards any prior distribution. This work selects a sample population ($\Nt{D}^{'}$) with 5k data points.
        \item \textbf{Random Pipeline Sampling}: This step random selects a subset of pipelines ($\Ns{P}^{'}$) from $\Ns{P}$ to be used for generating HoE. This work samples 10\% of pipelines as sampled pipeline population.
    \end{itemize}
    \item \textbf{Generating HoE}: To generate the HoE ($\Ng{H}$, a hashmap), one must iterate over each pipeline in the sub-sampled pipelines ($\Ns{P}^{'}$) and evaluate their performance against the representative data population ($\Ng{D}^{'}$). Along with pipeline performance, HoE records individual steps performance as well.
    
    Each pipeline $\Ng{P}\in \Ns{P}^{'}$, consists of several steps ($\Ns{S}$) which can be denoted as $\Ng{P} \leftarrow \langle\Ng{S}_1, \Ng{S}_2, ..., \Ng{S}_S\rangle$ where $\Ng{S}_i$ is the step $i$ and $S$ is the number of steps in the pipeline. For example, when $S=3$, the pipeline $\Ng{P}$ can be expressed as $\langle \Ng{S}_1,\Ng{S}_2,\Ng{S}_3\rangle$. To record the pipeline and its individual step performance in the $\Ng{H}$, the output of each data step ($\Ng{S}^{d}$) is passed through a surrogate model (as described in definition \ref{def:proposed:surrogate}) to extract the step accuracy, for model step ($\Ng{S}^{m}$), the trained model (of interest) accuracy on the sub-sampled data is recorded (i.e. $\Ng{D}^{'}$). Step-accuracy pair is interested in the hashmap, s.t. 
    \begin{equation}
        \Ng{H}[\Ng{S}] = 
        \begin{cases}
            \Ng{M}^{s}(\Ng{S}(\Ng{X}^{'}), \Ng{Y}^{'}) & \Ng{S}~\KwSty{is data step}\\
            \Ng{M}(\Ng{X}{'}, \Ng{Y}^{'}\vert \Ng{S}) & \Ng{S}~\KwSty{is model step}
        \end{cases}
    \end{equation}
    For example, of a 3-step pipeline, $p=\{s_1, s_2, s_3\}$, \algoabbr~records the performance of each step of this pipeline in $\Ng{H}$: $\langle\Ng{S}_1-acc_1\rangle$, $\langle\Ng{S}_1,\Ng{S}_2-acc_2\rangle$ and $\langle\Ng{S}_1,\Ng{S}_2,\Ng{S}_3-acc_3\rangle$, where $acc$ is the performance score (accuracy in the scope of classification task) evaluated after a step $\Ng{S}$ or multiple interlinked steps.
\end{enumerate}

\begin{algorithm}[t]
    \hspace{-2em} 
    \KwSty{evaluateStep}$(\Ng{S},\Nm{X},\Nm{Y})$: \\
        \KwData{Data and Pipeline Step}
        \KwResult{Same for output data}
        \eIf{$\Ng{S}$ is $\Ng{S}^{d}$}{
            $\Nm{X} \leftarrow \Ng{S}(\Nm{X})$ {\color{blue}\CommentSty{// Data Step}} \\
            $acc \leftarrow \Ng{M}^{s}(\Nm{X},\Nm{Y})$\\
        }{
            $acc \leftarrow \Ng{M} (\Nm{X}, \Nm{Y}\vert \Ng{S})$ {\color{blue}\CommentSty{// Model Step}} \\
        }
        \KwRet{$acc$, $\Nm{X}$}
    \vspace{0.5em}
    \caption{Algorithmic overview of Step execution and evaluation}
    \vspace{-0.5em}
    \label{algo:proposed:evaluateStep}
\end{algorithm}
\vspace{-0.4in}
\subsection{Pipeline Search}
Upon creation of the HoE using the approach described in the previous section, this section describes the pipeline search process that leverages an early termination criterion to automate the decision process of whether to train a pipeline from start to end or terminate at an intermediate step. During the pipeline search \algoabbr~frequently refers to $\Ng{H}$ in the decision making. Search involves:
\begin{enumerate}
    \item Pipeline search is an iterative process, where \algoabbr~iterates through every pipeline ($\Ng{P} \in \Ns{P}$) in the search space ($\Nt{P}$)and further iterates through each step ($\Ng{S} \in \Ng{P}$ and $\Ng{S} \in \Nt{S}$)in a pipeline.
    \item For any step, $\Ng{S}_s$, the search process first checks if the step exists in $\Ng{H}$ or not. This check is essential, as many AutoML systems execute steps that are redundant. If the step or the sequence of steps not previously seen is not present then the step(s) is evaluated else pass accuracy is leveraged.
    \item In the case, $\Ng{S}_s$ does not exist in the $\Ng{H}$, we leverage Algorithm \ref{algo:proposed:evaluateStep} that executes the individual step operation and also evaluates the step output using surrogate model (as described in Definition \ref{def:proposed:surrogate}).
    \item The resultant of the \textbf{pipeline evaluation} based on the surrogate model is an accurate estimate. This estimate is an approximation of how accurate the pipeline is. \algoabbr~operates under the assumption that ``\textit{intermediate step performance is reflective of the end pipeline performance}''. Therefore, the aim of \algoabbr~is to maximize the surrogate accuracy: i.e. 
    \vspace{-0.2cm}
    \begin{equation}
        \operatorname*{argmax}_{\forall \Ng{S}} \Ng{M}^{s}(\Ng{S}_{S-1}(...\Ng{S}_{1}(\Nm{X})),\Nm{Y})
    \end{equation}
    \item Once the accuracy is determined, we use the ``\textit{\textbf{Pipeline Terminate Criteria}}''. This criterion determines where the current step is sufficiently accurate enough to warrant further execution of the next steps or the pipelines, and is defined as follows:
    \vspace{-0.2cm}
    \begin{equation}
        earlyStop(\Ng{H}, \Ng{S}) = 
        \begin{cases}
            continue, & if~acc > median(\Ng{H}) \\
            terminate,& otherwise
        \end{cases}
    \end{equation}
    \item Winner pipeline is selected from the pipelines that finished until the last step and achieved maximum end accuracy.
\end{enumerate}

\noindent This concludes the proposed \algoabbr~framework. It is formally outlined the implementation details in Algorithm \ref{algo:proposed:hoe}. Next, we present experimental evaluation.

\begin{algorithm}[t]
     \KwData{Data ($\Nt{D} \leftarrow \langle \Nm{X}, \Nm{Y}\rangle$), and Pipelines ($\Ns{P}$)}
     \KwResult{Winner Pipeline}
     {\color{blue}\CommentSty{// Generating History of Experiments}}\\
     \KwSty{initialize:} $\Nt{D}^{'} \leftarrow represnetativeSampling(\Nt{D})$ \\
     \KwSty{initialize:} $\Ns{P}^{'} \leftarrow randomSampling(\Ns{P})$ \\
     \KwSty{initialize:} $ P \leftarrow \vert\Ns{P}^{'}\vert$\;
     $temp\Nm{X} \gets \Nm{X}^{'}$\;
     \For{$p\gets 1$ \KwTo $P$ \KwSty{by} $1$}{ 
       $\Ng{P}\gets \Ns{P}^{'}[p]$ \\
       $stepPrefix\gets  `` ''$\\
       \For{$\Ng{S} \in \Ng{P}$}{
        $stepPrefix \mathrel{+}= \Ng{S}$\;
        \If{$stepPrefix$ \KwSty{not in} $\Ng{H}$}{
            $acc,~temp\Nm{X} \leftarrow \KwSty{evaluateStep}(\Ng{S},\langle temp\Nm{X},\Nm{Y}\rangle)$\\
            $\Ng{H}[stepPrefix] \leftarrow acc$
        }    
       }
     }
     {\color{blue}\CommentSty{// Pipeline Search}}\\
     $temp\Nm{X} \gets \Nm{X}^{'}$\\
     \For{$\Ng{P} \in \{\Ng{P}_{1},\dots,\Ng{P}_{P}\}$}{
        $stepPrefix\gets  `` ''$\\
        \For{$\Ng{S} \in \Ng{P}$}{
            {\color{blue}\CommentSty{// Pipeline Evaluation Criterion}}\\
            $stepPrefix \mathrel{+}= \Ng{S}.prefix$\\
            \eIf{$\Ng{H}[stepPrefix]$}{
                $acc \leftarrow \Ng{H}[stepPrefix]$ {\color{blue}\CommentSty{// Identical Steps exists in $\Ng{H}$}} \\
            }{
                $acc,~temp\Nm{X} \leftarrow \KwSty{evaluateStep}(\Ng{S},\langle temp\Nm{X},\Nm{Y}\rangle)$\\
                $\Ng{H}[stepPrefix] \leftarrow acc$
            }    
            {\color{blue}\CommentSty{// Pipeline Termination Criterion}}\\
            \eIf{$acc > median(\Ng{H})$}{
                \KwSty{continue} {\color{OliveGreen}\CommentSty{// \textbf{CONTINUE} to next step in current pipeline}}
            }{
                \KwSty{break} {\color{red}\CommentSty{// \textbf{TERMINATE} current pipeline execution}}
            }
        }  
     }
     $winnerPipeLinePrefix$, $acc \leftarrow max(\Ng{H})$\;
     \KwRet{$winnerPipeLinePrefix$, $acc$}\;
     \vspace{0.2em}
     \caption{Algorithmic overview of the proposed framework that involves the generation of History of Experiments ($\Ng{H}$), evaluation of pipeline steps (i.e. use of $\Ng{M}^{s}$) and early termination of a pipeline which is driven by the recorded history of pipeline evaluations as the AutoML systems for optimal configuration of the pipeline.
     }
     \label{algo:proposed:hoe}
\end{algorithm}

\clearpage
\begin{table*}[t]
    \centering
    \caption{Overview of benchmark datasets}
    \label{tab:exp:datasets}
    \resizebox{1.0\linewidth}{!}{
      \begin{tabular}{|l|rrrrrrcc|}
	 	\hline
	 	\multirow{2}{*}{\textbf{Name}} & 
	 	\multirow{2}{*}{\textbf{Id}} & 
	 	\multirow{2}{*}{\textbf{Instances}} & 
	 	\multicolumn{3}{c}{\textbf{Number of Input Features}} & 
	 	\multirow{2}{*}{\textbf{Classes}} & 
	 	\textbf{Missing} & 
	 	\textbf{Imbalanced}\\ \cline{4-6}
	 	& & & \textbf{Numeric} & \textbf{Categorical} & \textbf{Total} & & 
	 	\textbf{Data} & \textbf{Data} \\ \hline 
	 	\Tstrut Adult Salary Prediction  & 1590  & 48842  & 6   & 8  & 14  & 2   & $\checkmark$ & $\checkmark$ \\
	 	Airlines Flight Delay            & 1169  & 539383 & 3   & 4  & 7   & 2   & --           & $\checkmark$ \\
	 	Amazon Employee Access           & 4135  & 32769  & 0   & 9  & 9   & 2   & --           & $\checkmark$ \\
	 	Australian Credit Approval       & 40981 & 690    & 6   & 8  & 14  & 2   & --           & $\checkmark$ \\
	 	Bank Marketing                   & 1461  & 45211  & 7   & 9  & 16  & 2   & --           & $\checkmark$ \\
	 	Blood Transfusion Service Center & 1464  & 748    & 4   & -- & 4   & 2   & --           & $\checkmark$ \\
	 	Car                              & 40975 & 1728   & 0   & 6  & 6   & 4   & --           & $\checkmark$ \\
	 	CMAE-9                           & 1468  & 1080   & 856 & -- & 856 & 9   & --           & --           \\
	 	Connect-4                        & 40668 & 67557  & 0   & 42 & 42  & 3   & --           & $\checkmark$ \\
	 	CoverType                        & 1596  & 581012 & 10  & 44 & 54  & 7   & --           & $\checkmark$ \\
	 	Credit-G                         & 31    & 1000   & 7   & 13 & 20  & 2   & --           & $\checkmark$ \\
	 	Fabert                           & 41164 & 8237   & 800 & -- & 800 & 7   & --           & $\checkmark$ \\
	 	Helena                           & 41169 & 65196  & 27  & -- & 27  & 100 & --           & $\checkmark$ \\
	 	Higgs                            & 23512 & 98050  & 28  & -- & 28  & 2   & $\checkmark$ & $\checkmark$ \\
	 	Jungle Chess                     & 41027 & 44819  & 6   & -- & 6   & 3   & --           & $\checkmark$ \\
	 	kc1                              & 1067  & 2109   & 21  & -- & 21  & 2   & --           & $\checkmark$ \\
	 	KDDCup09 Appetency               & 1111  & 50000  & 192 & 38 & 230 & 2   & $\checkmark$ & $\checkmark$ \\
	 	kr-vs-krp                        & 3     & 3196   & 0   & 36 & 36  & 2   & --           & $\checkmark$ \\
	 	MiniBooNE                        & 41150 & 130064 & 50  & -- & 50  & 2   & --           & $\checkmark$ \\
	 	Nomao                            & 1486  & 34465  & 89  & 29 & 118 & 2   & --           & $\checkmark$ \\
	 	NumerAI 28.6                     & 23517 & 96320  & 21  & -- & 21  & 2   & --           & $\checkmark$ \\
	 	Phoneme                          & 1489  & 5404   & 5   & -- & 5   & 2   & --           & $\checkmark$ \\
	 	Segmet                           & 40984 & 2310   & 19  & -- & 19  & 7   & --           & --           \\
	 	Shuttle                          & 40685 & 58000  & 9   & -- & 9   & 7   & --           & $\checkmark$ \\
	 	Sylvine                          & 41146 & 5124   & 20  & -- & 20  & 2   & --           & --           \\
	 	Vehicle                          & 54    & 846    & 18  & -- & 18  & 4   & --           & $\checkmark$ \\
	 	\hline
	  \end{tabular}   
	}
\end{table*}

\section{Experiments}\label{sec:experiments} 
In this section, we present in detail the experimental evaluation of the proposed \textit{\underline{\algoabbr~Framework}} and its performance on 26 benchmark datasets \cite{gijsbers2019open}, and is compared against state-of-the-art AutoML systems \cite{feurer2015efficient,olson2016tpot,mlbox2018romblay}.

\subsection{Configuration}\label{sec:experiments:config}
The experiments were performed on a server running Ubuntu 20.04 with AMB EPYC 7302 16-core processor with 64 threads, 512 GB RAM, and 4x NVIDIA RTX A6000 (48GB Memory) with Python version 3.8.10.

\subsection{Benchmark Datasets}\label{sec:experiments:datasets}
We selected 26 datasets from AutoML Benchmark~\cite{gijsbers2019open}.This benchmark consists of datasets that have a varying number of instances, features, and classes in the datasets with the presence of missing data and imbalance in a number of instances associated with classes, as shown in Table \ref{tab:exp:datasets}. 

\subsection{State-of-the-art Competitors}\label{sec:exp:competitors}
In this work, we considered 3 different AutoML systems:
\begin{enumerate}
    \item \textbf{AutoSkLearn}: unlike existing AutoML system, uses meta-learning to warm start hyper-parameter search, such as Bayesian optimization, by leveraging is an AutoML system that leverages Bayesian optimization and builds an ensemble of machine learning models that were considered by search process \cite{feurer2015efficient}.
    
    \item \textbf{TPOT}: is a tree-based pipeline optimization tool that automates the building of ML pipelines by combining a flexible expression tree representation of pipelines searched using genetic programming \cite{olson2016tpot}.
    
    \item \textbf{MLBox}: is an open-source sequential AutoML system that allows for fast and distributed data pre-processing, feature selection and bayesian optimization-based hyperparameter search \cite{mlbox2018romblay}.
\end{enumerate}
A description of the settings of each AutoML system is in the following subsection.

\begin{table}[t]
    \centering
    \caption{Comparison of different training constraints used in various competitors against \algoabbr}
    \label{tab:exp:constraints}
    \begin{tabular}{|l|l|c|c|c|}
        \hline 
        \multicolumn{2}{|l|}{\textbf{AutoML System}} & \textbf{Sample} & \textbf{Search} & \textbf{Time} \\
        \hline
        \multicolumn{2}{|l|}{\Tstrut\textbf{AutoSkLearn}} & $\checkmark$ & BO & $\checkmark$ \\
        \multicolumn{2}{|l|}{\textbf{TPOT}}        & $\checkmark$ & GP & $\checkmark$ \\ 
        \hline
        \multirow{4}{*}{\textbf{MLBox}} & \Tstrut\textbf{Base}  & --         & GS & -- \\
                                        & \textbf{GS}    & $\checkmark$ & GS & -- \\
                                        & \textbf{BO}    & $\checkmark$ & BO & -- \\
                                        & \textbf{TC}   & $\checkmark$ & GS & $\checkmark$ \\
        \hline
        \multicolumn{2}{|l|}{\Tstrut\textbf{\algoabbr}} & $\checkmark$ & GS-ET & $\checkmark$ \\ 
        \hline
    \end{tabular}
\end{table}

\subsection{Training Constraints}\label{sec:exp:constraint}
\algoabbr~introduces various constraints while running an AutoML system (which trains multiple machine learning pipelines). In particular, these constraints help reduce the number of computing resources needed to identify the near-optimal, if not optimal, machine learning pipeline that achieves the best model accuracy then most. Additionally, these constraints allow for a fair comparison between \algoabbr~and other competitors (see discussed in Section \ref{sec:exp:competitors}). We have shown how different AutoML systems and their variants satisfy the constraints (as described below) in Table \ref{tab:exp:constraints}. In the table, ($-$) represents \textit{not applicable}, and ($\checkmark$) applicable.

\subsubsection{Data Sampling}
\algoabbr~observes that the contrary to convention of using entire data to train and evaluate pipelines - which is costly in time and needs compute resources one can achieve similar results by subsampling the data such that the sub-population is representative of the overall population and as a positive consequence reducing AutoML training time. In this paper, \algoabbr~samples 5000 representative points from the data to train and evaluate AutoML pipelines.

\subsubsection{Hyper-Parameter Search}
There are different hyper-parameter search strategies, such as grid search (GS) and Bayesian optimization (BO), that have their own benefits. Both work with predefined hand-crafted hyper-parameter search space and differ in how the space is explored. The former explores each hyper-parameter configuration sequentially to identify the best (making a costly process), whereas the latter, randomly samples a few starting configurations and is followed by random sampling around the prior samples that converge the objective function. It is important that when AutoML systems are compared one does not have an unfair advantage between the intermediate choices they make, therefore, we investigate different scenarios of how different choices affect the model outcome.

\subsubsection{Training Time}
As noted in \cite{feurer2015efficient,olson2016tpot}, resources are finite and there is a limit on time. Further, as results show \algoabbr, achieve anywhere between 3x to 250x reduction in AutoML training - thus it is important to understand the implication of same time constraints on competitors as well.

\begin{table}[t]
    \centering
    \caption{Model performance comparison between SOTA AutoML System various MLBox + \algoabbr}
    \label{tab:exp:accuracy}
    {\setlength{\tabcolsep}{1em}
        \begin{tabular}{|l|r|r|}
            \hline
            \textbf{AutoML Systems} & \textbf{Accuracy (\%)} & \textbf{\algoabbr~Wins}\\ 
            \hline
            \Tstrut
            \textbf{AutoSkLearn}    & 81.54 & 17 \\
            \textbf{TPOT}           & 82.08 & 14 \\
            \textbf{MLBox-TC}       & 83.30 & 24 \\
            \hline
            \Tstrut
            \textbf{\algoabbr}      & 83.42 & -- \\
            \hline
        \end{tabular}
    }
    \end{table}

\begin{table}[t]
    \centering
    \caption{Comparison of different training constraints used in various competitors against \algoabbr}
    \label{tab:exp:metrics}
    \resizebox{\linewidth}{!}{
        {\setlength{\tabcolsep}{1em}
        \begin{tabular}{|l|l|r|r|r|r|r|}
            \hline 
            \multicolumn{2}{|l|}{\textbf{AutoML}} & 
            \multicolumn{1}{|c|}{\textbf{Training}} & 
            \multicolumn{1}{|c|}{\textbf{\# of Pipelines}} & 
            \multicolumn{1}{|c|}{\textbf{Accuracy}} & 
            \multicolumn{2}{|c|}{\textbf{\algoabbr}} \\ \cline{6-7}
            \multicolumn{2}{|l|}{\textbf{System}} & 
            \multicolumn{1}{|c|}{\textbf{Time (secs)}} & 
            \multicolumn{1}{|c|}{\textbf{Evaluated}} & 
            \multicolumn{1}{|c|}{\textbf{Percentage}} &
            \multicolumn{1}{|c|}{\textbf{Dataset Wins}} & 
            \multicolumn{1}{|c|}{\textbf{Time Gains}}\\
            \hline
            \multirow{4}{*}{\textbf{MLBox}} & \Tstrut\textbf{Base} & 28318.92 & 432 & 83.86 &  7 &  40x\\
                                            & \textbf{GS}          &  1804.65 & 432 & 83.42 & 23 &   3x\\
                                            & \textbf{BO}          &  1860.46 & 432 & 83.41 & 23 &   3x\\
                                            & \textbf{TC}          &   496.35 & 153 & 83.30 & 24 & Same\\
            \hline
            \multicolumn{2}{|l|}{\Tstrut\textbf{\algoabbr}}        &   496.35 & 116 & 83.42 & -- &   --\\ 
            \hline
        \end{tabular}
        }
    }
\end{table}

\subsection{Performance: \algoabbr~vs Competitors}\label{sec:exp:performance:competitors}
While different AutoML systems have different hyper-parameter search space and searching strategies, they may require different amounts of compute time. We apply the constraints listed in Table \ref{tab:exp:constraints} for a fair comparison. In particular, we limited the amount of time available for competitive AutoML systems to run until the wall clock time taken by \algoabbr. 

As shown in Table \ref{tab:exp:accuracy}, \algoabbr~is able to outperform the competitors when compared using aggregate performance metrics, i.e. the table reports average classification accuracy for all 26 benchmark datasets. We further present a win counter called ``\algoabbr~Wins''. This counter shows that for how many datasets \algoabbr~was able to win against the competitors.

We observe that the considerable gains observed in \algoabbr~can be credited to the early evaluation criteria that evaluated the data quality at a variety of intermediate steps in an AutoML system, contrary to the evaluation of the pipeline being performed at the last step.

\subsection{Performance: \algoabbr~vs Constraints}\label{sec:exp:performance:constraints}
In Table \ref{tab:exp:metrics}, we present the comparison of \algoabbr~against different constraints imposed on MLBox (we could not modify AutoSklearn and TPOT as they are shipped as end-to-end systems as opposed to modularized MLBox).

We observed that unconstrained MLBox(Base) outperforms \algoabbr~which is expected as it is a greedy search that considered the entire dataset rather than a sub-sampled data (as noted in \ref{sec:exp:constraint}), However, one must consider the trade-off along the time - while the accuracy gain is 0.44\% with MLBox(Base) the time taken to achieve that gain is 57x more than the time taken by \algoabbr. Furthermore, \algoabbr~is able to prune the search space to explore only 116 pipelines end-2-end as opposed to 432 in the MLBox(Base). 

When the data sampling constraint is imposed on MLBox(GS), there are no accuracy gains - both approaches achieve identical classification accuracy, however, MLBox(GS) requires 3.5x time to achieve the same level of accuracy. While both MLBox(Base) and (GS) use Grid Search which is not an intelligent searching technique, we also explore MLBox with Bayesian optimization (BO) to allow for intelligent traversal of search space to counter early termination of pipelines in \algoabbr~MLBox(BO) fails to beat \algoabbr~both in time and accuracy.
Furthermore, when both data sampling and time constraints are introduced simultaneously, MLBox(TC) suffers a loss in classification accuracy.

\begin{figure}[t]
    \centering
    \begin{subfigure}{0.48\linewidth}
        \centering
        \includegraphics[width=\linewidth]{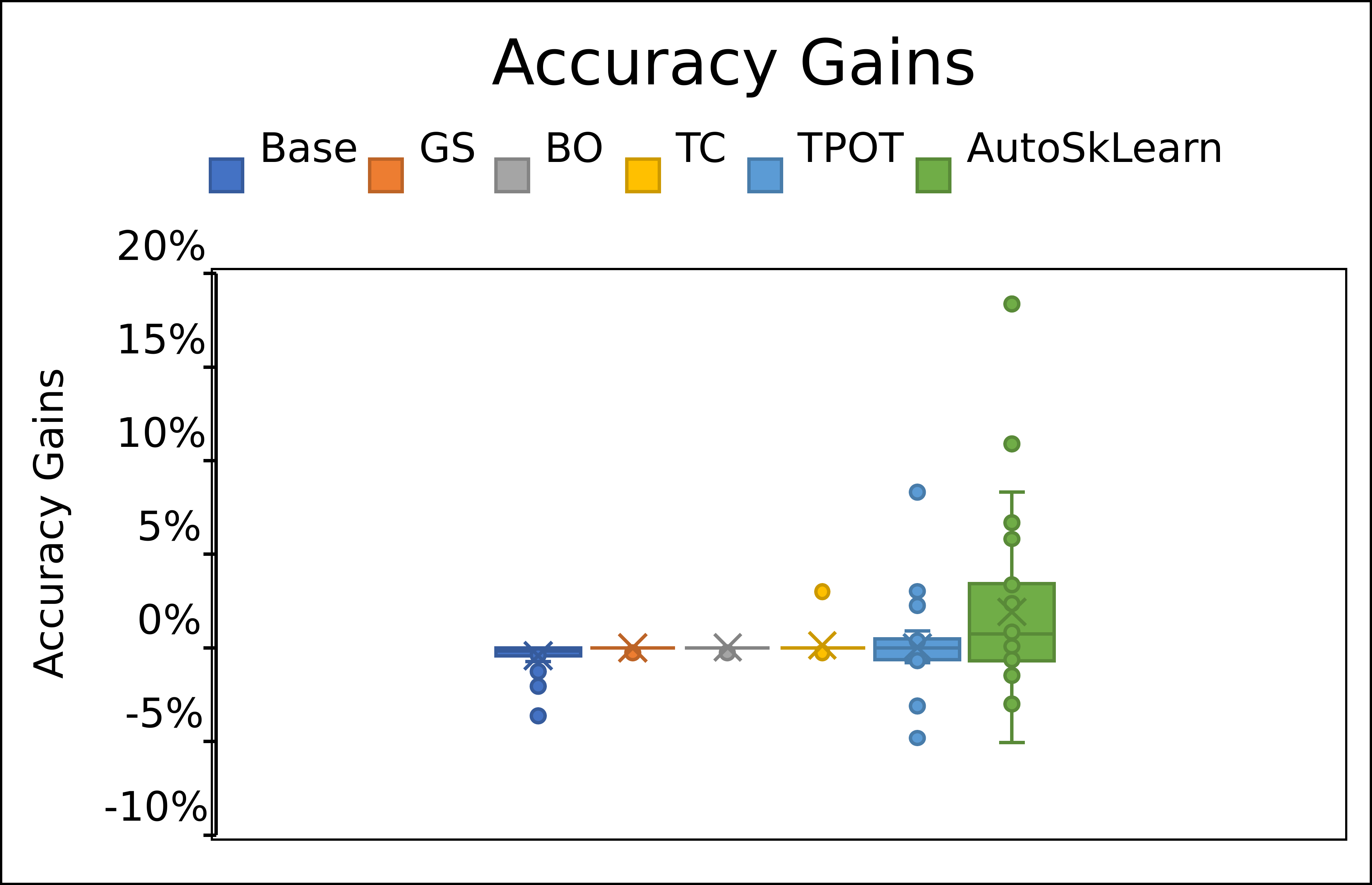}
        \caption{Accuracy Gains}
        \label{fig:exp:gains:accuracy}
    \end{subfigure}
    \begin{subfigure}{0.48\linewidth}
        \centering
        \includegraphics[width=\linewidth]{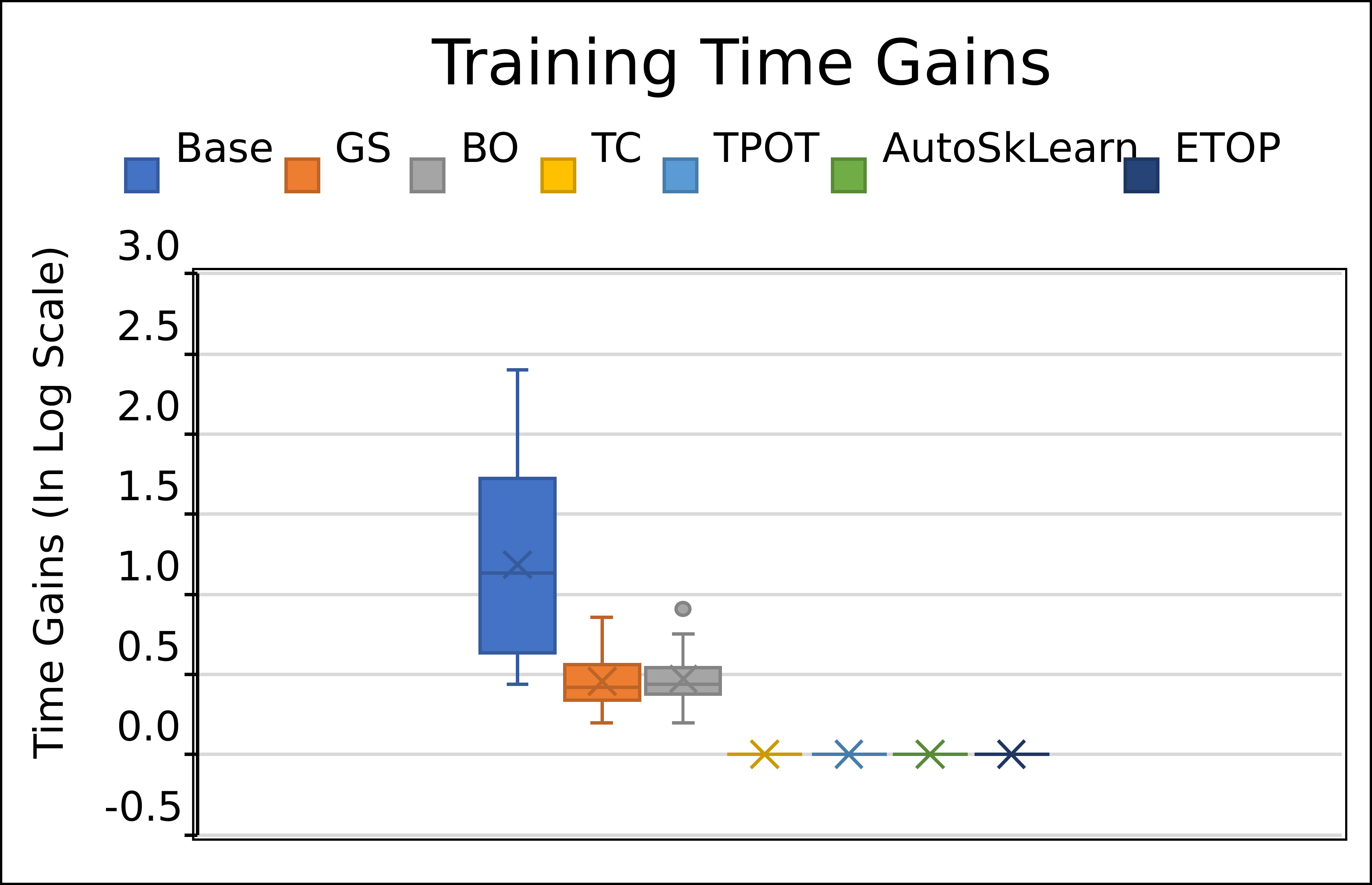}
        \caption{Training Time Gains}
        \label{fig:exp:gains:time}
    \end{subfigure}
    \caption{Comparison of accuracy and time gains observed for eTOP against competitors ($\uparrow$ -- higher the better)}
    \label{fig:exp:gains}
\end{figure}

\subsection{Gains observed with \algoabbr}\label{sec:exp:gains}
In Figures \ref{fig:exp:gains:accuracy} and \ref{fig:exp:gains:time}, we show the comparison of different AutoML systems (competitors) against \algoabbr, along the lines of gains observed for individual datasets. 
We define the accuracy gains ($\Ng{G}_A$) as:
\begin{equation}
    \Ng{G}_A=ACC_{\algoabbr} - ACC_{competitor},
\end{equation}
and, time gains ($\Ng{G}_T$) as:
\begin{equation}
    \Ng{G}_T=Time_{competitor}/Time_{\algoabbr}.
\end{equation}
Here, $Time$ is defined as the total time taken to traverse the search space to find the winning pipeline, and $ACC$ is defined as the end accuracy observed for the winning timeline.

\noindent The two Figures clearly show that the accuracy and time gains go hand in hand. For key competitors such as AutoSkLearn and TPOT, one can achieve comparable accuracy gains ($gains > 0$)when the systems are run for the same amount of time. However, for MLBOX-based models - which majorly rely on greedy search, leading to better (Base) or comparable performance (Gs/BO/TC) but at the cost of time. As the figures show, accuracy gains are in the negative, and time gains are in the positive.

\noindent In Table \ref{tab:dataset:performance} we present the dataset level performance for each competitor and \algoabbr. Furthermore, this table is used to derive the Tables \ref{tab:exp:accuracy} \& \ref{tab:exp:metrics}, and Figures \ref{fig:exp:gains:accuracy} \& \ref{fig:exp:gains:time}.

\begin{table*}[ht]
  \centering
  \caption{Detailed overview of observed accuracies and training times for all 26 datasets across different state-of-the-art competitors and variants of MLBox. ($-$ represents AutoML system was not able to run for the dataset)}
  \rotatebox{90}{
  \resizebox{1.5\linewidth}{!}{
    \begin{tabular}{|r|r|r|r|r|r|r|r|r|r|r|r|r|r|r|}
        \hline
        \multicolumn{1}{|c|}{\multirow{3}{*}{\textbf{Datasets}}} & \multicolumn{7}{c|}{\textbf{Model Accuracy (\%)}}          & \multicolumn{7}{c|}{\textbf{Model Training Time (\textit{in seconds})}} \\
    \cline{2-15}          & \multicolumn{4}{c|}{\Tstrut\textbf{MLBox}} & \multicolumn{1}{c|}{\multirow{2}{*}{\textbf{TPOT}}} & \multicolumn{1}{c|}{\multirow{2}{*}{\textbf{AutoSk}}} & \multicolumn{1}{c|}{\multirow{2}{*}{\textbf{ETOP}}} & \multicolumn{4}{c|}{\textbf{MLBox}} & \multicolumn{1}{c|}{\multirow{2}{*}{\textbf{TPOT}}} & \multicolumn{1}{c|}{\multirow{2}{*}{\textbf{AutoSk}}} & \multicolumn{1}{c|}{\multirow{2}{*}{\textbf{ETOP}}} \\
    \cline{2-5}\cline{9-12}          & \multicolumn{1}{c|}{\Tstrut\textbf{Base}} & \multicolumn{1}{c|}{\textbf{GS}} & \multicolumn{1}{c|}{\textbf{BO}} & \multicolumn{1}{c|}{\textbf{TC}} & \multicolumn{1}{c|}{} & \multicolumn{1}{c|}{} &       & \multicolumn{1}{c|}{\textbf{Base}} & \multicolumn{1}{c|}{\textbf{GS}} & \multicolumn{1}{c|}{\textbf{BO}} & \multicolumn{1}{c|}{\textbf{TC}} & \multicolumn{1}{c|}{} & \multicolumn{1}{c|}{} &  \\
        \hline
\textbf{1596} & 96.47\ & 96.47\ & 96.46\ & 96.46\ & 94.08\ & 78.08\ & 96.47\ & 264016.79 & 2302.49 & 2369.84 & 1059.88 & 1059.88 & 1059.88 & 1059.88 \\
        \textbf{1169} & 65.78\ & 64.01\ & 64.01\ & 64.01\ & -- & 63.00\ & 63.73\ & 56967.57 & 956.86 & 1448.22 & 604.14 & 604.14 & 604.14 & 604.14 \\
        \textbf{41150} & 93.52\ & 92.30\ & 92.30\ & 92.30\ & 89.23\ & 92.16\ & 92.25\ & 78598.07 & 2450.91 & 2013.65 & 594.05 & 594.05 & 594.05 & 594.05 \\
        \textbf{23512} & 71.71\ & 71.61\ & 71.52\ & 71.52\ & 72.21\ & 69.24\ & 71.61\ & 27120.65 & 1721.97 & 1639.62 & 534.64 & 534.64 & 534.64 & 534.64 \\
        \textbf{23517} & 52.16\ & 51.86\ & 51.86\ & 51.86\ & 52.55\ & 50.51\ & 51.86\ & 25762.82 & 1341.76 & 1208.80 & 499.38 & 499.38 & 499.38 & 499.38 \\
        \textbf{40668} & 82.26\ & 81.85\ & 81.85\ & 81.85\ & 82.01\ & 76.04\ & 81.85\ & 23569.78 & 1353.41 & 1670.67 & 413.31 & 413.31 & 413.31 & 413.31 \\
        \textbf{41169} & 35.10\ & 31.46\ & 31.46\ & 28.46\ & 27.98\ & 24.78\ & 31.46\ & 112071.83 & 8716.16 & 6697.97 & 1211.98 & 1211.98 & 1211.98 & 1211.98 \\
        \textbf{40685} & 99.99\ & 99.97\ & 99.83\ & 99.83\ & 99.99\ & 99.14\ & 99.97\ & 8902.93 & 860.63 & 558.58 & 356.62 & 356.62 & 356.62 & 356.62 \\
        \textbf{1111} & 98.20\ & 98.20\ & 98.20\ & 98.20\ & -- & 98.30\ & 98.20\ & 73037.93 & 5350.73 & 5496.79 & 970.70 & 970.70 & 970.70 & 970.70 \\
        \textbf{1590} & 86.53\ & 86.48\ & 86.48\ & 86.48\ & -- & 85.21\ & 86.48\ & 8778.81 & 959.41 & 1156.18 & 385.76 & 385.76 & 385.76 & 385.76 \\
        \textbf{1461} & 90.65\ & 90.32\ & 90.32\ & 90.32\ & -- & 89.84\ & 90.58\ & 7192.07 & 984.42 & 1053.63 & 388.75 & 388.75 & 388.75 & 388.75 \\
        \textbf{41027} & 82.16\ & 81.59\ & 81.59\ & 81.59\ & 86.42\ & 78.21\ & 81.59\ & 3277.91 & 673.88 & 878.64 & 317.85 & 317.85 & 317.85 & 317.85 \\
        \textbf{1486} & 96.25\ & 95.94\ & 95.94\ & 95.94\ & 96.27\ & 95.20\ & 95.94\ & 14809.66 & 1846.61 & 1675.52 & 491.57 & 491.57 & 491.57 & 491.57 \\
        \textbf{4135} & 94.48\ & 94.29\ & 94.29\ & 94.29\ & 94.85\ & 94.23\ & 94.29\ & 7364.02 & 1542.23 & 1834.83 & 412.00 & 412.00 & 412.00 & 412.00 \\
        \textbf{41164} & 68.05\ & 68.05\ & 68.05\ & 68.05\ & 71.17\ & 69.54\ & 68.05\ & 10491.37 & 7376.44 & 9955.23 & 1229.80 & 1229.80 & 1229.80 & 1229.80 \\
        \textbf{1489} & 86.70\ & 86.70\ & 86.70\ & 86.70\ & 91.21\ & 89.36\ & 86.70\ & 1237.61 & 725.10 & 862.94 & 317.74 & 317.74 & 317.74 & 317.74 \\
        \textbf{41146} & 93.44\ & 93.36\ & 93.36\ & 93.36\ & 94.24\ & 94.73\ & 93.44\ & 2233.91 & 1043.63 & 1052.97 & 457.23 & 457.23 & 457.23 & 457.23 \\
        \textbf{3} & 99.26\ & 99.18\ & 99.18\ & 99.18\ & -- & 99.84\ & 99.18\ & 1880.98 & 934.20 & 915.72 & 333.35 & 333.35 & 333.35 & 333.35 \\
        \textbf{40984} & 91.99\ & 91.23\ & 91.23\ & 91.23\ & 95.88\ & 96.32\ & 91.23\ & 1437.89 & 854.73 & 780.91 & 315.06 & 315.06 & 315.06 & 315.06 \\
        \textbf{1067} & 86.13\ & 85.83\ & 85.83\ & 85.83\ & 86.26\ & 86.73\ & 85.83\ & 1248.58 & 690.35 & 890.44 & 315.70 & 315.70 & 315.70 & 315.70 \\
        \textbf{40975} & 95.88\ & 95.66\ & 95.66\ & 95.66\ & -- & 96.24\ & 95.66\ & 1105.41 & 587.15 & 772.37 & 325.14 & 325.14 & 325.14 & 325.14 \\
        \textbf{1468} & 93.40\ & 93.40\ & 93.40\ & 93.40\ & 93.06\ & 93.06\ & 93.40\ & 1716.88 & 1398.59 & 809.59 & 389.51 & 389.51 & 389.51 & 389.51 \\
        \textbf{31} & 75.75\ & 75.50\ & 75.50\ & 75.50\ & -- & 78.50\ & 75.50\ & 833.40 & 644.13 & 846.04 & 231.97 & 231.97 & 231.97 & 231.97 \\
        \textbf{54} & 77.37\ & 77.37\ & 77.37\ & 77.37\ & 76.47\ & 66.47\ & 77.37\ & 1074.69 & 599.87 & 381.38 & 239.80 & 239.80 & 239.80 & 239.80 \\
        \textbf{1464} & 78.26\ & 78.26\ & 78.26\ & 78.26\ & 76.00\ & 74.67\ & 78.26\ & 637.13 & 471.51 & 660.22 & 232.87 & 232.87 & 232.87 & 232.87 \\
        \textbf{40981} & 88.77\ & 88.04\ & 87.86\ & 88.04\ & 79.71\ & 79.71\ & 88.04\ & 923.21 & 533.71 & 741.11 & 274.40 & 274.40 & 274.40 & 274.40 \\
        \hline
        \multicolumn{15}{c}{}\\
        \hline
        \textbf{Average} & 83.86 & 83.42 & 83.41 & 83.30 & 82.08 & 81.54 & 83.42 & 28318.92 & 1804.65 & 1860.46 & 496.35 & 496.35 & 496.35 & 496.35 \\ \hline
    \end{tabular}%
  }
  }
  \label{tab:dataset:performance}%
\end{table*}

\section{Conclusion}\label{sec:conclusion}
In this paper, we address the problem of prohibitive computations cost of executing AutoML systems that stems from end-to-end execution of AutoML pipelines. We propose the \textit{\underline{\algoabbr~Framework}} that significantly reduces the execution time of the AutoML systems by at least 3x and at best 40x while maintaining comparable accuracy results. \algoabbr~achieve the said gains by using an intelligent early termination criterion that self-determines if the pipeline in an AutoML system should be terminated at any intermediate step as opposed to being trained until the end. Experimental results on 26 benchmark datasets show wide gains observed using the proposed approach from small to large datasets.

\newpage
\bibliographystyle{splncs04}
\bibliography{references}

\end{document}